\documentclass[10pt,journal,compsoc]{IEEEtran}
\usepackage{amsmath,amsfonts}
\usepackage{algorithmic}
\usepackage{algorithm}
\usepackage{array}
\usepackage[caption=false,font=normalsize,labelfont=sf,textfont=sf]{subfig}
\usepackage{textcomp}
\usepackage{stfloats}
\usepackage{url}
\usepackage{verbatim}
\usepackage{graphicx}
\usepackage{cite}
\usepackage{multirow}
\usepackage{multicol}
\usepackage{xspace}
\usepackage[most]{tcolorbox}
\usepackage{booktabs}
\newcommand{\myparagraph}[1]{\vspace{1em}\noindent\textbf{#1}\hspace{1em}}
\newcommand{\our}{\textsf{MACR}\xspace}
\begin{document}

\title{Navigating Unreliable Parametric and Contextual Knowledge: Explicit Knowledge Conflict Resolution for LLM Inference}

\author{Huang Peng, Jiuyang Tang, Weixin Zeng, Hao Xu, Xiang Zhao
\thanks{Huang Peng, Jiuyang Tang, Weixin Zeng, Hao Xu and Xiang Zhao are with the National Key Laboratory of Big Data and Decision, National University of Defense Technology, China.}
\thanks{Manuscript received April 19, 2021; revised August 16, 2021.}}

\markboth{Journal of \LaTeX\ Class Files,~Vol.~14, No.~8, August~2021}%
{Peng \MakeLowercase{\textit{et al.}}: Navigating Unreliable Parametric and Contextual Knowledge: Explicit Knowledge Conflict Resolution for LLM Inference}

\IEEEpubid{0000--0000/00\$00.00~\copyright~2021 IEEE}

\maketitle

\begin{abstract}
Large language models (LLMs) have achieved strong performance across a wide range of language-based tasks by leveraging both extensive parametric knowledge and in-context learning ability, enabling them to incorporate external information provided in the input prompt. However, the integration of external knowledge can introduce conflicts, not only between the model’s internal parametric knowledge and the external information, but also among multiple pieces of external contexts. 
Existing approaches typically assume that either the model or the provided context is reliable, overlooking the possibility that both sources may contain errors, and avoid conflicts by privileging one source over the other, rather than actively resolving inconsistencies. To address these limitations, we propose a novel framework \our for LLM knowledge conflict resolution that \emph{moves beyond the conventional binary choice paradigm and incorporates an explicit conflict-resolution mechanism based on a multi-agent reasoning approach}. 
Specifically, we first propose an \emph{adaptive knowledge assessment and retrieval} approach that employs a modified semantic entropy measure to quantify an LLM’s confidence in its answer to a given query. Based on this confidence estimation, \our either externalizes the model’s internal knowledge as textual representations or retrieves relevant external knowledge when internal knowledge is insufficient, generating basic contexts for subsequent reasoning. Then we introduce an \emph{inductive multi‑agent reasoning framework} with three specialized agents that, respectively, induce explicit rules, analyze potential conflicts, and resolve inconsistencies across all available contexts. Empirical results demonstrate that \our significantly outperforms state-of-the-art baselines across benchmarks, while also providing interpretable resolutions of explicit conflicts.
\end{abstract}


\section{Introduction}
\IEEEPARstart{L}{arge} Language Models (LLMs) have demonstrated remarkable capabilities, leveraging their extensive parametric knowledge acquired during pre-training to perform a wide array of language-based tasks. A key factor contributing to their success is the ability for in-context learning, which allows LLMs to dynamically incorporate and reason over information provided within the input prompt. This has given rise to powerful paradigms such as Retrieval-Augmented Generation (RAG)~\cite{DBLP:conf/nips/LewisPPPKGKLYR020}, where external knowledge is retrieved and supplied as context to enhance the model's responses, granting it a broader knowledge scope and improving the factual accuracy of its outputs.

However, the integration of external knowledge introduces a critical challenge: the emergence of knowledge conflicts~\cite{DBLP:journals/corr/abs-2310-00935}. Such conflicts arise when information contained within the retrieved context contradicts the LLM's internal, parametric knowledge~\cite{DBLP:conf/emnlp/XuQGW0ZX24}.
The root of these discrepancies is multifaceted; they typically stem from the LLM possessing outdated knowledge~\cite{DBLP:conf/naacl/LuuKGMS22,DBLP:journals/tacl/DhingraCEGEC22}, or from the presence of inconsistent information within the external context itself~\cite{DBLP:conf/ijcnlp/PanCKW23}. 
When these conflicts occur, regardless of the source of the error, the model's performance can be significantly degraded, leading to the generation of factually incorrect or misleading answers~\cite{DBLP:conf/emnlp/PhamNLN24}. 
Therefore, resolving these knowledge conflicts by discerning the reliable information—whether from the context or the model's internal beliefs—is paramount to ensuring the trustworthiness of LLM-generated content. 
The central problem this paper addresses is \emph{how to enable an LLM to navigate these conflicts, critically evaluate both its internal knowledge and the provided context, and synthesize a response that mitigates the influence of any factual errors, irrespective of their origin}.

\begin{figure}
    \centering
    \includegraphics[width=\linewidth]{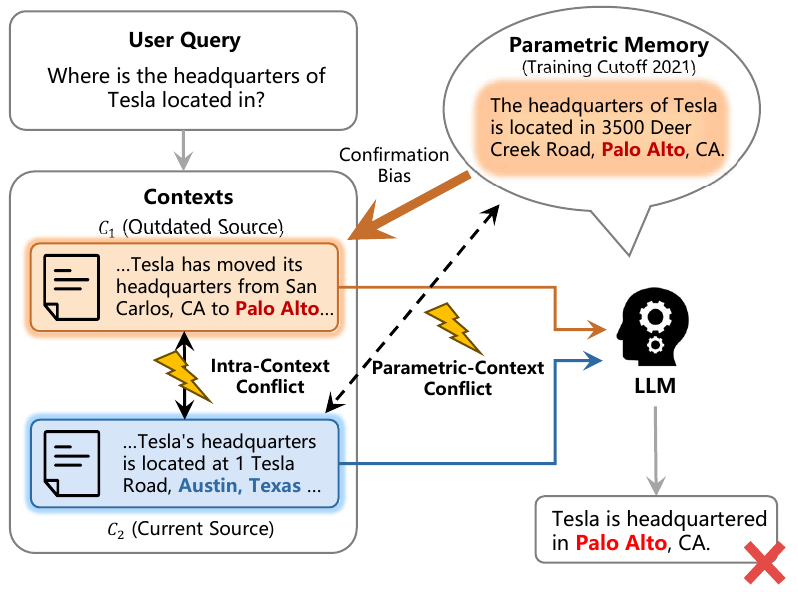}
    \caption{An example of knowledge conflict in retrieval-augmented generation. The model faces dual conflicts: conflicts among external contexts and contradictions between parametric memory and external context. Because the outdated context ($C_1$) corroborates the model's internal memory, the LLM produces an erroneous answer.}
    \label{fig1}
\end{figure}

Prior efforts towards this problem mainly adopt a restrictive conflict-avoidance paradigm based on binary source selection~\cite{floridi2023ai}. 
By presuming that either the LLM’s parametric knowledge or the retrieved context is correct~\cite{DBLP:conf/emnlp/XuQGW0ZX24}, these methods \emph{collapse the inherently complex task of conflict resolution into a simplistic choice of allegiance}.  
Even more recent dynamic strategies~\cite{DBLP:conf/emnlp/KimKPPCKYLK24,DBLP:journals/corr/abs-2503-15888}, which switch between sources using LLM confidence scores, remain constrained by this same “either/or” assumption. This paradigm is fundamentally limited because it sidesteps explicit conflict analysis. In realistic settings, such as the example in Figure~\ref{fig1}, retrieved contexts may contain internal inconsistencies, and the wrong information may align with outdated parametric knowledge, reinforcing confirmation bias. 
Existing methods that merely pick a single ``trusted'' source thus fail to interrogate underlying discrepancies and cannot robustly handle cases where both sources are unreliable. Consequently, when the chosen source is flawed, these systems tend to propagate errors rather than resolve them and cannot provide explicit, interpretable conflict analysis and resolution.

To address these limitations, we propose a novel framework \our for LLM knowledge conflict resolution that moves beyond the binary choice paradigm and conducts explicit conflict resolution. 
Our approach is built upon two key innovations: an \textbf{adaptive knowledge assessment and retrieval} approach and an \textbf{inductive multi-agent reasoning} process. Firstly, our framework employs an \textbf{adaptive knowledge assessment and retrieval} to ensure a balanced and robust evaluation of information without assumptions that prefer a certain source. The process begins by assessing the LLM's confidence in its own knowledge using semantic entropy, a step designed to detect uncertainty and mitigate the influence of hallucinations. Crucially, rather than keeping internal knowledge implicit, we explicitly articulate it in textual form or retrieve pertinent information from external sources. This explicit representation enables direct comparison between internal knowledge and diverse external contexts, thereby supporting unified conflict detection and resolution. Moreover, when the model lacks sufficient prior knowledge, external information can serve as an anchor that stabilizes its reasoning and guides it toward more reliable outputs. Secondly, we introduce an \textbf{inductive multi-agent reasoning} paradigm. This approach orchestrates three distinct LLM agents: an \textit{Observer}, an \textit{Analyzer}, and a \textit{Reasoner}. The \textit{Observer} operates on the training data to inductively summarize and extract generalizable rules for resolving knowledge conflicts. During the inference phase, the \textit{Analyzer} is responsible for identifying and breaking down the specific discrepancies among all contexts including internal knowledge of the LLM or retrieved external information, while the \textit{Reasoner} applies the rules learned by the \textit{Observer} to systematically resolve these conflicts and synthesize the final answer.

In summary, the contributions of this paper are threefold:
\begin{itemize}
    \item We propose a novel conflict resolution framework \our that can effectively handle the complex and realistic scenario where both the LLM's internal knowledge and the retrieved external context may be erroneous or inconsistent; 
    \item We introduce two methodological innovations: an \textbf{adaptive knowledge assessment and retrieval} that externalizes internal knowledge or retrieve external information, and an \textbf{inductive multi-agent reasoning} strategy that utilizes specialized agents, i.e., Observer, Analyzer, Reasoner, to systematically learn rules and resolve conflicts.
    \item We demonstrate through empirical evaluation that our method significantly improves the accuracy and reliability of LLM responses in the presence of conflicting and potentially faulty information.
\end{itemize}

\section{Related Works}
Our research is situated at the intersection of two critical areas in contemporary large language model research: knowledge conflict resolution and robust retrieval-augmented generation. This section reviews seminal and recent works in both domains, contextualizing our contributions within the existing literature.

\subsection{Knowledge Conflict Resolution in Large Language Models}
The integration of external information into LLM prompts is a standard practice for enhancing response quality. However, this frequently leads to conflicts between the model's pre-trained parametric knowledge and the provided non-parametric, contextual knowledge. A significant body of work has emerged to address this challenge, largely bifurcating into two opposing philosophical stances.

The first line of research operates under the assumption that the provided context is the authoritative source of truth. Methods in this category aim to steer the model's output to align more closely with the contextual information, effectively suppressing its internal knowledge when a discrepancy is detected~\cite{DBLP:conf/acl/0002HHLPL22,DBLP:conf/emnlp/GekhmanHAES23,DBLP:conf/emnlp/ZhouZPC23,DBLP:conf/emnlp/XueW0M0WSJ0W23}. For example, KAFT~\cite{DBLP:conf/acl/LiRZWLVYK23} involves constructing training data and fine-tuning the model to ensure it adheres to relevant documents while disregarding irrelevant ones. Contrastive Decoding (CAD)~\cite{DBLP:conf/naacl/ShiHLTZY24} amplifies the probability of tokens that are more likely given the context-aware model compared to a context-free baseline. Similarly, IRCAN~\cite{DBLP:conf/nips/ShiJSDWX24} proposes an inference-time method that calibrates model outputs to increase faithfulness to the provided context. These approaches are effective when the external knowledge is reliable, but they risk forcing the model to repeat factual errors present in the context.

Conversely, the second line of research posits that an LLM’s intrinsic parametric knowledge is often more reliable than potentially noisy external contexts. Accordingly, these methods prioritize resilience against contextual perturbations, utilizing training or prompting techniques that encourage the model to disregard erroneous external information in favor of its internal memory~\cite{DBLP:conf/eacl/WellerKWLD24,DBLP:conf/acl/XuLYZS0FX024}. For instance, Pan et al.~\cite{DBLP:conf/emnlp/PanPCNKW23} propose three defense strategies—misinformation detection, prompting, and a divide-and-vote mechanism—to mitigate the influence of noisy external data. Similarly, Xu et al.~\cite{DBLP:conf/acl/XuLYZS0FX024} propose a prompt-based approach that inserts a dedicated system prompt to caution the LLM about potential misinformation and to instruct it to verify its memorized knowledge before generating a response. Adopting a detection-based approach, Hong et al.~\cite{DBLP:conf/naacl/HongKKMW24} train a discriminator on a dataset of counterfactual noise; upon identifying noisy documents, the system inserts a specific identifier into the prompt to explicitly alert the model to the presence of misinformation.

Recognizing the limitations of a fixed-stance approach, more recent studies have explored dynamic or adaptive strategies. For instance, Li et al.\cite{DBLP:journals/corr/abs-2503-10996} introduced a framework capable of switching between context-following and knowledge-recalling modes, though it requires a manual, pre-specified setting. A more sophisticated dynamic approach was proposed by~\cite{DBLP:journals/corr/abs-2503-15888} and~\cite{DBLP:conf/emnlp/KimKPPCKYLK24}, which use the model's output confidence score as a heuristic to decide whether to trust the internal parametric knowledge or the external context on a per-query basis. If the model exhibits high confidence in its own answer, it relies on its internal knowledge; otherwise, it defaults to the provided context. While these methods represent a step forward, their core logic remains a binary choice: they dynamically select one source to trust, thereby avoiding the conflict rather than resolving it. A crucial limitation is their inability to handle the prevalent and complex scenario where both the internal knowledge and the external context are fallible. Our work directly addresses this gap by proposing a mechanism to evaluate both sources concurrently and reason through their discrepancies.

\subsection{Robustness in Retrieval-Augmented Generation}
Retrieval-Augmented Generation has become a cornerstone for building knowledge-intensive LLM applications. The standard RAG pipeline involves retrieving relevant documents from an external corpus and prepending them to the user's query as context for the LLM. The quality of the retrieved documents is paramount; however, real-world retrieval systems are imperfect and often return documents containing noise~\cite{DBLP:conf/sigir/CuconasuTSFCMTS24,DBLP:conf/iir/CuconasuTSFCMTS24}.

In the context of RAG, "noise" has traditionally been defined as information that is irrelevant or tangential to the user's query~\cite{DBLP:conf/acl/ZhangXXWLWS25,DBLP:journals/corr/abs-2511-10375}. A prominent body of work focuses on identifying and mitigating this form of noise. For example, Corrective RAG (CRAG)~\cite{DBLP:journals/corr/abs-2401-15884} introduces a lightweight retrieval evaluator to assess the overall quality of retrieved documents for a given query. If the documents are deemed relevant, a knowledge refinement step extracts the most salient information. If they are irrelevant or ambiguous, CRAG supplements the process with web search to find better information. Similarly, InstructRAG~\cite{DBLP:conf/iclr/WeiC025} fine-tunes an LLM to proactively detect noise within the context and generate a "critical analysis" of the provided documents before synthesizing a final answer. TruthfulRAG~\cite{DBLP:conf/aaai/LiuSZ26} additionally constructs a knowledge graph (KG) from the provided contexts and leverages this structure to facilitate reasoning and resolve conflicting information.

While these methods enhance robustness against irrelevance, they pay less attention to the problem of factual conflicts, where a retrieved document is relevant but factually incorrect. A recent work, CARE-RAG~\cite{DBLP:journals/corr/abs-2507-01281}, moves closer to our research problem. It first allows the LLM to generate an initial response based on its internal knowledge. It then explicitly compares this initial output with the retrieved contextual documents, performs a conflict analysis, and finally instructs the LLM to generate a revised answer based on both knowledge sources and the analysis. While CARE-RAG also externalizes internal knowledge and contrasts it with contextual evidence, it does not explicitly account for hallucinations or gaps in the model’s knowledge, limiting its ability to handle cases where neither source is fully reliable.

Our work builds on insights from these advanced RAG frameworks but differs in its explicit and interpretable mechanism for resolving conflicts. In particular, we account for the potential unreliability of LLM outputs through a dedicated assessment component, and we design a rule-based, multi-agent reasoning framework that yields more comprehensive and well-justified analyses of both conflicts and their synthesized resolutions.

\section{Methodology}
This section details our proposed framework for knowledge conflict resolution, designed to navigate and resolve discrepancies between a LLM internal knowledge and external context. Our method is structured into three main stages: Knowledge Assessment, Knowledge Retrieval, and Inductive Multi-Agent Reasoning.

\subsection{Problem Formulation}
The core challenge we address is enabling a LLM to navigate and resolve conflicts between its internal, parameterized knowledge and external, contextual knowledge to produce a factually accurate answer. We formalize this problem as follows.

Let a user's query be represented by a question that seeks to identify an entity, $t^*$, which holds a specific relation, $r$, with a subject entity, $h$, regardless to the format or expression form of the query. We assume for simplicity that this query has a single correct answer, forming the knowledge triple $(h, r, t^*)$.

The LLM possesses its own internal, parametric knowledge, which may or may not be correct or complete. We can model this internal knowledge regarding the query as a triple $(h, r, t_0)$, where $t_0$ is the tail entity the model "believes" to be correct. It is also possible that the model has no specific knowledge for this triple, in which case $t_0$ is null.

Simultaneously, the model is provided with an external context, $C$, composed of a set of $n$ textual statements or documents. Each statement, $c_{(h,r,t_i)}$, contains information pertaining to the query, asserting that the entity $t_i$ is the answer. This can be represented as a set of contextual knowledge triples: $C = \{c_{(h,r,t_i)}\}_{i=1}^n$.

A knowledge conflict occurs when the entities proposed by these different sources are not identical, i.e., when $t_0 \neq t_i$ for some $i$, or when $t_i \neq t_j$ for some $i \neq j$. The sources of these conflicts are varied: the model's internal knowledge $(h, r, t_0)$ may be outdated or incorrect, and any of the contextual statements $c_{(h,r,t_i)}$ could contain misinformation, be out of date, or be misinterpreted. 
The task of knowledge conflict resolution, therefore, is to design a function, $f$, that takes the user's query $(h, r, ?)$, the model's internal knowledge $(h, r, t_0)$, and the set of external contexts $C$ as input, and outputs the correct entity $t^*$:
\begin{equation}
    t^* = f((h, r, ?), (h, r, t_0), C).
\end{equation}

To achieve this, the function $f$ must effectively perform a credibility assessment, discerning the correct knowledge while identifying and mitigating the influence of all other incorrect knowledge triples. The ultimate goal is to generate a final answer that is grounded exclusively in the identified correct knowledge.
\begin{figure*}
    \centering
    \includegraphics[width=0.7\linewidth]{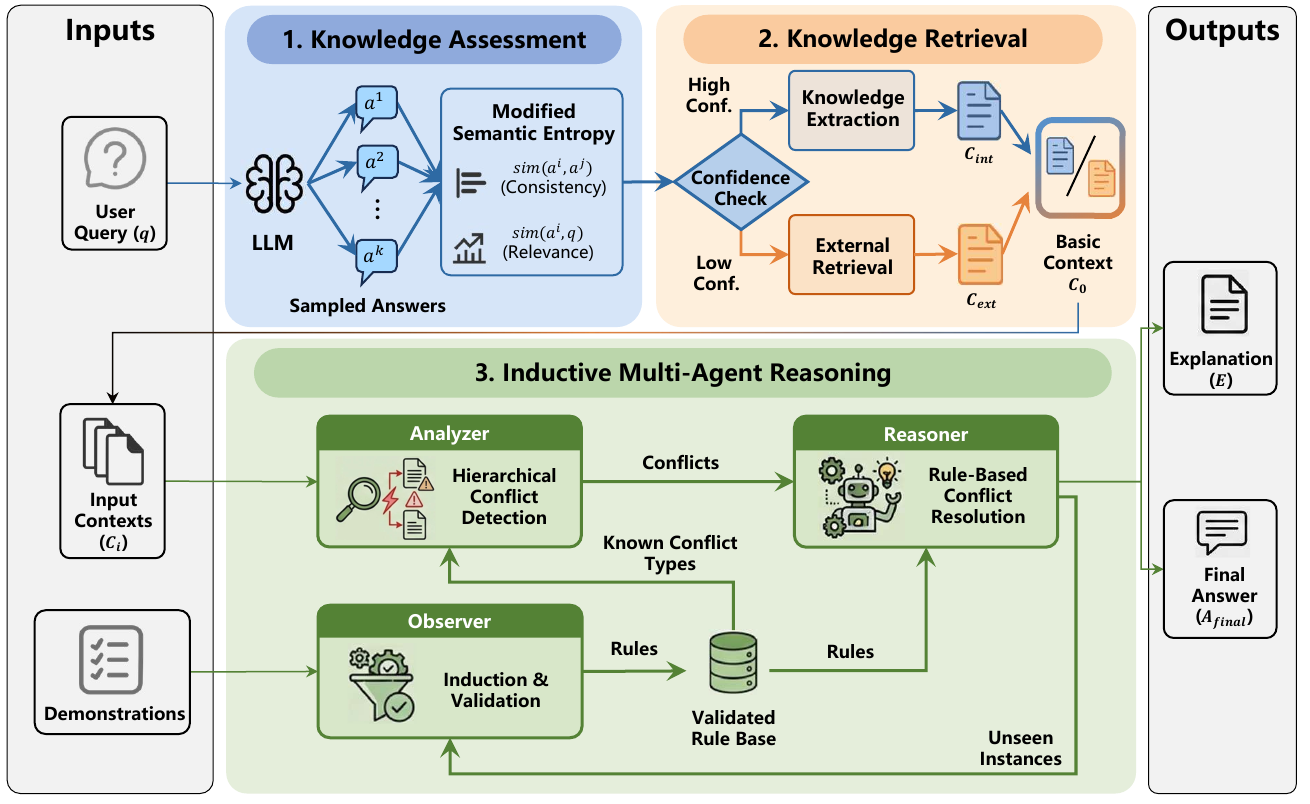}
    \caption{The framework of proposed method.}
    \label{fig2}
\end{figure*}

\subsection{Framework Overview}

Our proposed framework is structured into three sequential modules designed to systematically evaluate, retrieve, and reconcile knowledge for robust reasoning: \textbf{Knowledge Assessment}, \textbf{Knowledge Retrieval}, and \textbf{Inductive Multi-Agent Reasoning}. 
The process begins with the Knowledge Assessment module evaluates the LLM's uncertainty regarding the input query. To detect hallucinations and overconfidence, we employ a modified semantic entropy metric that weighs the consistency among sampled answers, i.e., $sim(a^i,a^j)$, and their relevance to the question, i.e., $sim(a^i, q)$. 
Based on this assessment, the Knowledge Retrieval module executes a branching strategy: high confidence triggers the externalization of the model's parametric knowledge into a tangible reasoning path ($C_{int}$), while low confidence activates the retrieval of reliable external information ($C_{ext}$). Either internal or external knowledge will serve as the basic context ($C_0$) in the next module. 
Finally, the Inductive Multi-Agent Reasoning module resolves conflicts between this basic context ($C_0$) and contexts from input ($C_i$). This is achieved through a collaborative triad of agents: an \textit{Observer} that first inducts conflict resolution rules from training data and rigorously validates and filters them against a held-out set based on coverage and support metrics to ensure a high-quality rule base; an \textit{Analyzer} that performs hierarchical conflict detection across context pairs during inference; and a \textit{Reasoner} that applies the Observer's validated rules to the Analyzer's findings to synthesize a conflict-free final answer ($A_{final}$) and a explanation ($E$).

\subsection{Knowledge Assessment}
The initial and critical step of our framework is to ascertain the LLM's inherent confidence regarding a given query, $q$. A reliable confidence metric allows the system to determine whether to primarily trust the LLM's internal knowledge or to seek external validation. We propose a multi-faceted confidence measure that integrates semantic consistency with contextual metadata.

The foundation of our metric is semantic entropy~\cite{DBLP:journals/nature/FarquharKKG24}. Unlike conventional approaches that depend on token-level probabilities which can be unstable, we assess the semantic diversity among multiple candidate responses generated by the LLM for a given query $q$. To this end, we employ the improved Semantic Neighborhood Neighborhood Entropy (SNNE) method, which enhances computational efficiency by omitting clustering and boosting steps~\cite{DBLP:conf/acl/NguyenPM25}. Specifically, we generate $k$ distinct outputs, denoted by $\{a^1, a^2, ..., a^k\}$, using stochastic decoding strategies such as top-p sampling. Then, the semantic entropy for a given query, $H_q$, is then given by:
\begin{equation}
    H(q) = -\sum_{i=1}^{k} P(a^i|q) \log \left(\sum_{j=1}^k \exp\left(f(a^i, a^j)\right)\right),
\end{equation}\label{eq.1}
where $f(\cdot,\cdot)$ is a similarity function between two answers, $P(a^i|q)$ is the probability that the LLM generate answer $a^i$ to query $q$. A higher value of $H_q$ reflects greater semantic divergence among candidate answers, suggesting lower model confidence in its responses.

Semantic entropy quantifies uncertainty by measuring the dispersion of the LLM's generated responses; however, it overlooks the intrinsic content and relevance of the answers. For instance, if a model consistently generates uninformative responses such as ``I don't know'', the calculated entropy would be deceptively low despite the lack of actual knowledge. To address this limitation, we introduce a modification to the semantic entropy calculation. We incorporate the semantic similarity between the answer and the question, denoted as $\text{sim}(a_i, q)$, into the computation of the pairwise answer similarity $\text{sim}(a_i, a_j)$. Formally, the modified semantic entropy is calculated as:
\begin{equation}
    H_{sem}(q) = -\sum_{i=1}^{k} P(a^i|q) \log \left(\sum_{j=1}^k \exp\left(f(a^i, a^j, q)\right)\right),
\end{equation}\label{eq.2}
where $f(a^i,a^j,q)=sim(a^i,a^j)\cdot sim(a^i,q)$. Consequently, the modified semantic entropy yields a low value (indicating high confidence) only when the model's responses are both internally consistent and semantically relevant to the input query.

In addition, several factors can distort this confidence measurement, including randomness in decoding, temporal misalignment, and subject entity ambiguity. While the process of aggregating multiple responses inherently mitigates randomness, the latter two factors necessitate explicit intervention. To address these sources of error, we augment our confidence estimation by explicitly incorporating two informational signals into the LLM's prompt:

(1) \textbf{Temporal Information} ($I_{\text{time}}$): Because LLMs are trained on static corpora, they may provide outdated information with unwarranted confidence. To mitigate this, we include up-to-date temporal context in the prompt, for example: ``The current date is October 2025. Please note that your training data may not reflect recent events,'' thereby explicitly signaling that the model’s knowledge may be obsolete.

(2) \textbf{Subject Disambiguation} ($I_{\text{subject}}$): Ambiguity in entity references—for instance, between ``Apple'' the company and ``apple'' the fruit—can yield erroneous confidence estimates. To address this, we append concise disambiguating information about the intended entity, such as key attributes or a brief description.

By proactively embedding these contextual signals within the prompt, the measurement of the model's confidence becomes more robust against temporal and subject-based confounders. The resulting confidence score,
\begin{equation}
    C(q) = H_{sem}(I_{\text{time}}\oplus I_{\text{subject}}\oplus q),
\end{equation}
where entropy is calculated over context-enriched response generations, serves as the decisive criterion for downstream processing in our framework.

\subsection{Knowledge Retrieval}
\label{sec:knowledge_retrieval}

This module functions as a conditional branch determined by the confidence score $C(q)$ derived from the Knowledge Assessment stage. We utilize a pre-defined threshold, $\tau$, to direct the system effectively: scores below $\tau$ indicate high model confidence and therefore prompt reliance on parametric knowledge, whereas scores above $\tau$ signal low confidence and consequently trigger external retrieval.

\myparagraph{Case 1: High Confidence ($C(q) < \tau$)}
When the LLM exhibits high confidence, we posit that its parametric knowledge is likely sufficient. To fully leverage this, we employ a structured, two-step generation process to externalize implicit beliefs into explicit text.

(1) \textbf{Generate Internal Context ($C_{int}$):} We first prompt the LLM, denoted by $\mathcal{M}$, to externalize its latent knowledge about the query $q$. To this end, we design an instruction set $\mathcal{I}_{ctx}$ that elicits the model’s internal knowledge, yielding:
\begin{equation}
    C_{int} = \mathcal{M}(q, \mathcal{I}_{ctx}).
\end{equation}

Concretely, the model is asked to produce: key attributes of the main subject entity in $q$; domain-relevant background information; ancillary details needed for comprehensive understanding; and cut-off time of its trained knowledge. This structured decomposition ensures that $C_{int}$ provides a rich, multifaceted representation of the model’s internal stance.

(2) \textbf{Generate Initial Answer ($A_{init}$):} Subsequently, the model formulates a direct answer, $A_{init}$, strictly conditioned on the generated internal context $C_{int}$. This ensures logical consistency between the evidence and the answer:
\begin{equation}
    A_{init} = \mathcal{M}(q, C_{int}, \mathcal{I}_{ans}),
\end{equation}
where $\mathcal{I}_{ans}$ denotes the instruction to reason based on the provided evidence.

\myparagraph{Case 2: Low Confidence ($C(q) \ge \tau$)}
Conversely, a low confidence score signals potential knowledge gaps, hallucinations, or outdated information. Here, the framework bypasses internal parameters and activates a retrieval mechanism $R$ to fetch evidence from an external corpus $\mathcal{D}_{ext}$.

(1) \textbf{Retrieve External Context ($C_{ext}$):} The query $q$ is utilized to retrieve raw information $\mathcal{D}_{raw} = R(q, \mathcal{D}_{ext})$. To handle diverse data formats, we select the most pertinent information as follows: for unstructured data, the retrieved content is segmented into knowledge chunks $\mathcal{S} = \{c_1, c_2, \dots, c_m\}$. We identify the definitive external context $C_{ext}$ by maximizing semantic similarity with the query:
\begin{equation}
    C_{ext} = \operatorname*{argmax}_{c_k \in \mathcal{S}} \operatorname{sim}(c_k, q),
\end{equation}
where $\operatorname{sim}(\cdot)$ denotes the cosine similarity between the chunk and query embeddings. For structured data, this corresponds to extracting the most relevant one-hop entity-relation subgraph.

(2) \textbf{Generate Initial Answer ($A_{init}$):} Similar to the branch of high confidence, the LLM then generates the initial answer $A_{init}$, conditioned solely on the reliable external evidence $C_{ext}$. Notably, since the low confidence score signals that the model's internal knowledge is likely absent or unreliable, this strategy inherently avoids conflicts and minimizes the risk of hallucination from parametric memory.
\begin{equation}
    A_{init} = \mathcal{M}(q, C_{ext}, \mathcal{I}_{ext}),
\end{equation}
where $\mathcal{I}_{ext}$ instructs the model to synthesize the answer strictly from the retrieved information.

The final output of this module is a unified tuple $\mathcal{O}_{ret} = (A_{init}, C_0)$, where $C_0$ represents the supporting information, either the internal context $C_{int}$ from the high-confidence branch or the external context $C_{ext}$ from the low-confidence branch. This tuple is subsequently passed to the next module for conflict analysis.

\subsection{Inductive Multi-Agent Reasoning}
\label{sec:multi_agent_reasoning}

This final module addresses the complexity of resolving contradictions between the model's (or retrieved) evidence ($C_0$) and various user-provided contexts ($C_1, \dots, C_n$), as well as inconsistencies among the contexts themselves. Mirroring the ``observe-conclude-analyze-apply-reason'' cycle in human cognitive process, we propose a framework orchestrated by three specialized agents: the \textbf{Observer}, \textbf{Analyzer}, and \textbf{Reasoner}. These agents collaborate to induce rules from data and apply them to resolve conflicts during the inference stage.

\subsubsection{Agent Definitions}\hfill

\myparagraph{The Observer ($\mathcal{A}_{obs}$)}
The Observer operates primarily during the offline training phase. Its role is to identify generalizable patterns of knowledge conflicts (e.g., temporal discrepancies, source reliability issues) from datasets. It is responsible for generating a candidate rule base and, crucially, validating these rules against a held-out set to ensure high quality before deployment.

\myparagraph{The Analyzer ($\mathcal{A}_{ana}$)} 
The Analyzer functions during the online inference phase. It is tasked with a hierarchical conflict detection process. Instead of comparing raw texts directly, it first synthesizes individual answers for each context to identify contradictions at the semantic level. Upon detecting a conflict, it drills down into the source texts to extract the precise conflicting snippets and categorizes the conflict type (e.g., ``Temporal Conflict'').

\myparagraph{The Reasoner ($\mathcal{A}_{reas}$)} 
The Reasoner acts as the final decision-maker. It synthesizes the final output by applying the Observer's validated rules to the Analyzer's specific findings. It retrieves rules relevant to the identified conflict type, verifies if the textual conditions are met, and applies the rule's resolution logic to arbitrate the correct information, ultimately generating the final answer and an explanation.

\subsubsection{Conflict Resolution Workflow}

The framework operates in two distinct phases: Rule Induction (offline) and Conflict Resolution Inference (online).

\myparagraph{Phase 1: Rule Induction with Verification.}
In this phase, the Observer constructs a robust knowledge base for conflict resolution using a demonstration set $\mathcal{D}_{demo}$ and a validation set $\mathcal{D}_{val}$.

(1) \textbf{Rule Generation:} The Observer scans $\mathcal{D}_{demo}$ to extract a set of candidate rules. Each rule $r$ is structured as a tuple: $r = \langle \text{Type}, \text{Condition}, \text{Resolution} \rangle$.
    
(2) \textbf{Rule Verification:} To filter out low-quality or hallucinated rules, we evaluate each candidate on $\mathcal{D}_{val}$ using two metrics:
\begin{itemize}
    \item \textbf{Coverage ($Cov$):} The frequency with which a rule's \textit{Type} and \textit{Condition} are met in the validation set.
    \item \textbf{Support ($Sup$):} The frequency with which applying the rule's \textit{Resolution} yields the correct ground-truth answer.
\end{itemize}

The final Rule Base $\mathcal{R}_{final}$ retains only those rules that exceed predefined thresholds $\delta_{cov}$ and $\delta_{sup}$:
\begin{equation}
    \mathcal{R}_{final} = \{ r \in \mathcal{R}_{cand} \mid Cov(r) \ge \delta_{cov} \land Sup(r) \ge \delta_{sup} \}
\end{equation}

\myparagraph{Phase 2: Conflict Resolution Inference.}
During inference, given a query $q$ and the set of available contexts $\{C_0, C_1, \dots, C_n\}$ (where $C_0$ is the output of the Knowledge Retrieval module), the system proceeds as follows:

(1) \textbf{Answer-Level Conflict Detection:} First, the Analyzer prompts the LLM to generate an independent answer $A_i$ for each context $C_i$. It then performs pairwise comparisons between all answers $(A_i, A_j)$.
    
(2) \textbf{Snippet Extraction:} If a semantic contradiction is found between $A_i$ and $A_j$, the Analyzer examines the source contexts $C_i$ and $C_j$. It extracts the specific conflicting text snippets $c_{i,k} \in C_i$ and $c_{j,l} \in C_j$, and identifies the conflict type $T$. This forms a conflict triplet:
\begin{equation}
    \psi = \langle c_{i,k}, c_{j,l}, T \rangle.
\end{equation}
    
(3) \textbf{Rule-Based Resolution:} For each conflict triplet $\psi$, the Reasoner queries $\mathcal{R}_{final}$ for rules matching type $T$. It checks if the snippets satisfy the rule's \textit{Condition} (the rule body). The rule's \textit{Resolution} (the rule head) is then applied to determine the credible information, yielding a local judgment $v_\psi$.

(4) \textbf{Update Feedback Loop:} If the Reasoner cannot find any rule $r \in \mathcal{R}_{final}$ such that $\psi$ satisfies the rule's \textit{Condition}, the system triggers a fallback mechanism. The Reasoner attempts to resolve the conflict utilizing the LLM's inherent parametric knowledge, producing a provisional judgment $v'_{fallback}$. Simultaneously, the conflict instance $\psi$ and the provisional judgment are transmitted back to the Observer. The Observer uses this feedback to update and refine $\mathcal{R}_{final}$ for future iterations, enabling the system to adapt to uncovered conflict patterns.
        
(5) \textbf{Final Synthesis:} Finally, the Reasoner aggregates the set of context $C$ and the set of all local judgments $\mathcal{V} = \{v_\psi\}$ to synthesize the final answer $A_{final}$ and the explanatory summary $E$:
\begin{equation}
    A_{final}, E = \mathcal{M}(\mathcal{V}, C, q)
\end{equation}

\section{Experiments}
\subsection{Experimental Setup}
To empirically validate the effectiveness of \our, we conduct a series of experiments designed to assess its performance in identifying and resolving knowledge conflicts. This section outlines the experimental setup, including the benchmark datasets, evaluation metrics, baseline models, and implementation details.

\myparagraph{Benchmarks.} We evaluate our method on three distinct datasets, each designed to test an LLM's behavior in the presence of conflicting or counterfactual information.
\begin{itemize}
    \item \textsf{ConflictBank}~\cite{DBLP:conf/nips/SuZQ0LSLZC24} is a comprehensive dataset specifically curated for knowledge conflict scenarios. For each query, it provides three types of conflicting contexts: misinformation conflicts, where the context contains factually incorrect statements; temporal conflicts, where the context presents information with temporal drift; and semantic conflicts, where the context contains ambiguous subject with query. Its diverse conflict types make it an ideal benchmark for testing the robustness of our framework.
    \item \textsf{ConFiQA}~\cite{DBLP:conf/acl/BiH0YZHMFLWDSZL25} originally developed for evaluating the contextual faithfulness of LLMs in RAG settings, ConFiQA contains questions paired with counterfactual contexts. The primary challenge is for the model to avoid being misled by the plausible but incorrect information provided in the prompt. We use this dataset to measure our model's ability to identify and discard faulty external knowledge.
    \item \textsf{MQuAKE}~\cite{DBLP:conf/emnlp/ZhongWMPC23} is multi-hop question-answering dataset that primarily used in the context of knowledge editing to evaluate whether a model's parametric knowledge has been successfully updated. It features questions where the correct answer requires knowledge that contradicts the model's pre-trained beliefs. We leverage MQuAKE to assess our framework's capacity to handle multi-hop knowledge conflicts.
\end{itemize}

\myparagraph{Metrics.} We employ two standard metrics Exact Match (EM) and ROUGE-L score in question-answering to evaluate the quality and accuracy of the generated answers.

\myparagraph{Baselines.} We compare our method against a suite of baseline models that represent current approaches to handling knowledge from different sources:
\begin{itemize}
    \item Direct Answer (\textsf{Direct}): The answer of LLM to given question without instruction and additional context, reflecting the interior knowledge of LLM.
    \item In-Context Learning (\textsf{ICL}): Given all context to the LLM including noisy information, reflecting the basic ability of LLM that confront noise.
    \item Robust RAG Method (\textsf{InstructRAG}~\cite{DBLP:conf/iclr/WeiC025}, \textsf{TruthfulRAG}~\cite{DBLP:conf/aaai/LiuSZ26}): We also implement two robust RAG methods \textsf{InstructRAG} and \textsf{TruthfulRAG}, which utilizes the ability of LLM to identify noise among contexts. This baseline represents the method that consider multiple context pieces and resolve conflicts among them. We reproduce the in-context learning version of \textsf{InstructRAG} that denoted as $\textsf{InstructRAG}_{\textsf{ICL}}$.
    \item Dynamic Selection Method (\textsf{CK-PLUG}~\cite{DBLP:journals/corr/abs-2503-15888}): We implement a representative dynamic method \textsf{CK-PLUG} that uses the LLM's output confidence as a heuristic to choose between its parametrically-generated answer and a context-based answer. This baseline reflects the current state-of-the-art that circumvents conflicts by selecting one source over the other.
\end{itemize}

\myparagraph{Implementation Details.} Due to resource constraints, we randomly sampled 1,000 instances from the dataset and split them into demonstration, validation, and test sets with a 1:2:7 ratio. 
In the Knowledge Assessment module, we calculated semantic entropy over $k=8$ generated samples, using the ROUGE-L score as the similarity function $f(\cdot,\cdot)$. The threshold $\tau$ was determined on the validation set as the maximum entropy value at which the model still answers correctly. In the Knowledge Retrieval module, we used GPT-4o-mini to obtain external knowledge when needed. For the Inductive Multi-Agent Reasoning module, we set the hyperparameters to $\delta_{cov}=0.05$ and $\delta_{sup}=0.6$. 
For all baselines, we used their publicly released code and ran them under the same hardware setup. To ensure a fair comparison, we also provided the same external knowledge generated by GPT-4o-mini to these baseline models.

\begin{table*}[htbp]
    \centering
    \caption{Comparison of overall performance across ConflictBank, ConFiQA, and MQuAKE benchmarks. The best results are \textbf{bolded}, and the second-best results are \underline{underlined}.}\label{overallres}
    \begin{tabular}{clcccccc}
        \toprule
        \multicolumn{1}{c}{\multirow{2}{*}{Model}} &\multicolumn{1}{c}{\multirow{2}{*}{Method}} & \multicolumn{2}{c}{\textsf{ConflictBank}} & \multicolumn{2}{c}{\textsf{ConFiQA}} & \multicolumn{2}{c}{\textsf{MQuAKE}} \\ 
        \cmidrule{3-8} & & EM & ROUGE-L & EM & ROUGE-L & EM & ROUGE-L \\
        \midrule
        \multirow{5}{*}{Llama3.1-8B} & \textsf{Direct} & 0.043 & 0.140 & 0.275 & 0.434 & 0.208 & 0.258 \\
        & \textsf{ICL} & 0.192 & 0.398 & 0.488 & 0.606 & 0.573 & 0.616 \\
        & $\textsf{InstructRAG}_{\textsf{ICL}}$ & \underline{0.312} & \underline{0.506} & 0.527 & 0.640 & 0.597 & 0.648 \\
        & $\textsf{TruthfulRAG}$ & 0.301 & 0.416 & 0.532 & 0.628 & 0.568 & 0.605 \\
        & \textsf{CK-PLUG} & 0.288 & 0.449 & \underline{0.544} & \underline{0.669} & \underline{0.634} & \underline{0.691} \\
        & \our & \textbf{0.549} & \textbf{0.678} & \textbf{0.750} & \textbf{0.832} & \textbf{0.920} & \textbf{0.924} \\
        \midrule
        \multirow{5}{*}{Qwen2.5-7B} & \textsf{Direct} & 0.022 & 0.092 & 0.190 & 0.322 & 0.259 & 0.328 \\
        & \textsf{ICL} & 0.302 & 0.456 & 0.466 & 0.616 & \underline{0.591} & 0.628 \\
        & $\textsf{InstructRAG}_{\textsf{ICL}}$ & 0.231 & 0.408 & 0.475 & 0.618 & 0.574 & 0.633 \\
        & $\textsf{TruthfulRAG}$ & 0.276 & 0.395 & \underline{0.558} & \underline{0.654} & 0.524 & 0.558 \\
        & \textsf{CK-PLUG} & \underline{0.326} & \underline{0.485} & 0.402 & 0.590 & 0.582 & \underline{0.647} \\
        & \our & \textbf{0.550} & \textbf{0.750} & \textbf{0.780} & \textbf{0.857} & \textbf{0.861} & \textbf{0.876} \\ \bottomrule
    \end{tabular}
\end{table*}

\subsection{Main Results}
This section presents the primary results of our empirical evaluation across the three benchmark datasets, containing two contradict context for each query. We analyze the performance of our proposed framework against the established baselines, focusing on the EM and ROUGE-L scores to demonstrate our method's superior capability in resolving knowledge conflicts. The main results are summarized in Table~\ref{overallres}.

From the performance results, we draw three main conclusions. (1) \our consistently achieves the best performance across all datasets, demonstrating its effectiveness in simultaneously resolving conflicts between external context and the model’s internal knowledge, as well as conflicts among multiple context pieces. (2) The performance of the $\textsf{Direct}$ baseline indicates that the LLM’s internal knowledge is highly deficient or inaccurate on $\textsf{ConflictBank}$. Consequently, all evaluated methods perform poorest on $\textsf{ConflictBank}$ while exhibiting relatively stronger performance on $\textsf{ConFiQA}$ and $\textsf{MQuAKE}$. (3) The two RAG methods $\textsf{InstructRAG}_{\textsf{ICL}}$ and \textsf{TruthfulRAG} yield highly comparable performances in most scenarios. This parity arises because both methods focus primarily on mitigating inter-contextual noise while neglecting the conflict between external contexts and internal priors. In contrast, while $\textsf{CK-PLUG}$ achieves competitive performance due to its ability to dynamically integrate internal and external knowledge, it fails to effectively resolve scenarios involving multiple contradictory contexts, ultimately leading to suboptimal utilization of external information.

\subsection{Performance on Variant Context}
To further investigate the robustness of our framework, we conduct an ablation study focused on the influence of context quantity and the proportion of noisy information.

\myparagraph{Experimental Design.}
We adapt the ConflictBank dataset for this study. For each query, we construct input prompts containing a varying number of contextual paragraphs. We ensure that exactly one of these paragraphs is factually correct, while the others are deliberately conflicting or erroneous. We vary the total number of contexts from 2 to 4, corresponding to noise ratios from 50\% to 75\%, thereby systematically increasing both the overall information content and the noise-to-signal ratio.

\begin{figure}[t]
    \centering
    \includegraphics[width=0.8\linewidth]{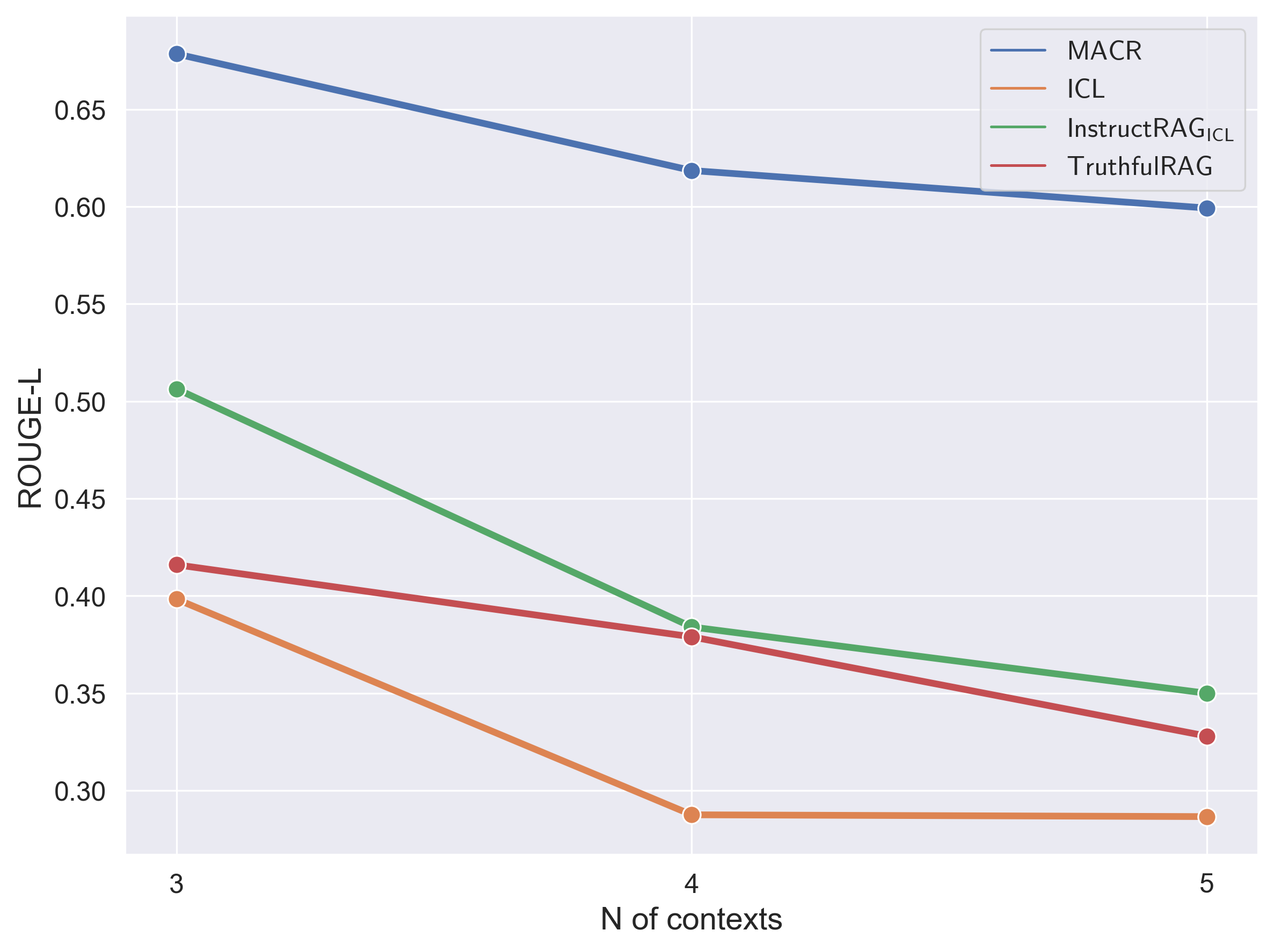}
    \caption{The degradation of ROUGE-L scores for each model as the number of conflicting contexts increases from 3 to 5.}
    \label{lchart}
\end{figure}

We evaluate our full framework and the two baselines, \textsf{ICL} and $\textsf{InstructRAG}_{\textsf{ICL}}$, under these increasingly challenging conditions. The results, presented in Table~\ref{vcontext} and visualized in Figure~\ref{lchart}, measure the models' ability to maintain accuracy as the contextual noise intensifies.

\begin{table}[t]
    \centering
    \caption{Performance under variant contexts on \textsf{ConflictBank}.}
    \label{vcontext}
    \resizebox{\linewidth}{!}{
    \begin{tabular}{lcccccc}
        \toprule
        \multicolumn{1}{c}{\multirow{2}{*}{Method}} & \multicolumn{2}{c}{$N=3$} & \multicolumn{2}{c}{$N=4$} & \multicolumn{2}{c}{$N=5$} \\
        \cmidrule{2-7}
        & EM & ROUGE-L & EM & ROUGE-L & EM & ROUGE-L \\
        \midrule
        \textsf{ICL} & 0.192 & 0.398 & 0.107 & 0.288 & 0.109 & 0.287 \\
        $\textsf{InstructRAG}_{\textsf{ICL}}$ & 0.312 & 0.506 & 0.223 & 0.384 & 0.256 & 0.350 \\
        $\textsf{TruthfulRAG}$ & 0.301 & 0.416 & 0.265 & 0.379 & 0.200 & 0.328\\
        \our & \textbf{0.549} & \textbf{0.678} & \textbf{0.470} & \textbf{0.549} & \textbf{0.350} & \textbf{0.599} \\
        \bottomrule
    \end{tabular}
    }
\end{table}

\myparagraph{Analysis of Results.}
Table \ref{vcontext} presents the performance evaluation on the \textsf{ConflictBank} dataset with varying numbers of contexts ($N \in \{3, 4, 5\}$). In this setup, increasing $N$ corresponds to introducing additional conflicting or noisy information, thereby elevating the difficulty of the task. 
The results reveal a clear distinction in robustness between the methods. The standard \textsf{ICL} baseline struggles significantly as the noise ratio increases, with its EM score plummeting from 0.192 at $N=3$ to roughly 0.10 at $N=4$ and 5, indicating a high susceptibility to distraction by contradictory contexts. While $\textsf{InstructRAG}_{\textsf{ICL}}$ and $\textsf{TruthfulRAG}$ shows marginal stability, its overall accuracy remains low, failing to effectively resolve the conflicts.
In contrast, \our shows markedly greater robustness. While its performance also decreases as the task becomes more complex, it consistently maintains a substantial margin over all baselines. Specifically, in terms of ROUGE-L, \our exhibits only a moderate decline (from 0.678 to 0.599), which represents the smallest performance drop among all compared methods. Notably, our method's performance in the most challenging setting ($N=5$) still significantly outperforms the best results achieved by the baselines in the easiest setting ($N=3$). This validates that our inductive multi-agent reasoning strategy effectively filters noise and identifies credible information, even when reliable evidence is heavily diluted by conflicting contexts.

\subsection{Ablation study}
To dissect the contribution of each key component within our proposed framework, we conduct a comprehensive ablation study. We design two distinct variants of \our by systematically removing the Knowledge Assessment and Retrieval module (w/o KA\&R), and replace the Inductive Multi-Agent Reasoning Module to a Chain-of-Thought (CoT) reasoning module (w/ CoT). By comparing the performance of these ablated versions against our full framework on the challenging \textsf{ConflictBank} dataset, we can quantify the impact of each component. The results are presented in Table \ref{tab:ablation}.

\myparagraph{Effectiveness of Knowledge Assessment and Retrieval.}
The variant \textit{w/o Assessment \& Retrieval} removes the initial confidence check, forcing the model to generate relevant information. We observe a distinct decline in the EM score for this variant. This drop indicates the critical necessity of our adaptive knowledge assessment and retrieval: without first assessing the LLM's internal uncertainty, the system fails to identify knowledge gaps. Consequently, it loses the ability to adaptively retrieve reliable external information when the model's parametric knowledge is absent or hallucinatory, leading to lower overall accuracy.

\myparagraph{Effectiveness of Inductive Multi-Agent Reasoning.}
The variant \textit{w/ CoT} replaces our specialized multi-agent module with a standard CoT prompting strategy. Specifically, the variant uses a single LLM to identify conflicts between $C_0$ and each $C_i$, then resolves them sequentially. The performance of this variant deteriorates sharply compared to the full model. This significant gap demonstrates that generic reasoning prompts are insufficient for handling complex knowledge conflicts. Our Inductive Multi-Agent approach provides a stable resolution mechanism; by explicitly applying induced rules rather than relying on unstructured generation, the multi-agent system ensures a far more robust and consistent adjudication between conflicting information sources.
\begin{table}[h!]
\centering
\caption{Ablation study results on the \textsf{ConflictBank} dataset. Each row represents the performance of our framework with a specific component removed or replaced.}
\label{tab:ablation}
\begin{tabular}{@{}lcc@{}}
\toprule
Model Variant & EM & ROUGE-L \\
\midrule
\our (Full) & \textbf{0.549} & \textbf{0.678} \\
\midrule
\textit{-w/o KA\&R} & 0.480 & 0.656 \\
\textit{-w/ CoT} & 0.229 & 0.329 \\
\bottomrule
\end{tabular}
\end{table}

\subsection{Case study}
\label{sec:case_study}

To provide intuitive insights into the internal mechanisms of \our, we analyze a specific case involving a temporal knowledge conflict. This scenario highlights how our \textbf{adaptive assessment and retrieval} and \textbf{inductive multi-agent reasoning} module effectively overrides the model's outdated parametric knowledge.

\subsubsection{Scenario Setup}\hfill

\textbf{Query ($q$):} ``\textit{What city is the headquarters of Tesla located in?}''

\textbf{Parametric Knowledge ($C_{int}$):} The base LLM (trained prior to late 2021) strongly associates Tesla's HQ with \textit{Palo Alto, California}.

\textbf{Input Context ($C_i$):} Two retrieved and conflicting snippets state: ``\textit{Tesla has moved its headquarters from San Carlos, CA to Palo Alto.}'' and ``\textit{Tesla's headquarters is located at 1 Tesla Road, Austin, Texas since 2021.}''

\subsubsection{Baseline Performance (Vanilla ICL)}
The baseline model receives two contradictory snippets simultaneously. Guided by a strong prior and insensitive to the temporal nuance of the query, it exhibits confirmation bias and prefers ``Palo Alto'' responding: ``\textit{Tesla is headquartered in Palo Alto, though it has operations in Austin.}''

\subsubsection{Analysis of \our}
Our framework processes this query through its three distinct stages, as illustrated in Table \ref{tab:case_study_process}:

(1) \textbf{Knowledge Assessment \& Retrieval:} The model initially displays high confidence in its internal knowledge. Following the instruction, it explicitly externalizes this belief into the basic context ($C_0$), providing detailed information, including the cutoff date of its training corpus: “\textit{Tesla is an American automotive, energy storage and solar power company ... as of 2021, the headquarters of Tesla is located in 3500 Deer Creek Road, Palo Alto, California.}” This explicit articulation is crucial because it converts implicit biases into analyzable text, thereby exposing latent assumptions that the LLM does not recognize as potentially incorrect.

(2) \textbf{Analyzer Agent (Conflict Detection):} The Analyzer first generates answers $A_i$ based on the corresponding contexts $C_i$: $A_0$ and $A_1$ both support ``Palo Alto'', whereas $A_2$ argues ``Austin''. It then scans all answer pairs $(A_i, A_j)$ and explicitly flags two contradictions, namely between $(A_0, A_2)$ and $(A_1, A_2)$.
    
(3) \textbf{Reasoner Agent (Rule Application):} The Reasoner receives the conflict flags and gathers the corresponding $C_i$ for each conflicting pair. Rather than resolving conflicts using its own priors, it queries the Observer’s rule base and matches each conflict to a specific \textbf{Temporal Update Rule}:
\begin{quote}
    \textit{Rule \#12:} Type: Temporal Conflict; Condition: A discrepancy exists between the dates of two statements; Resolution: The statement associated with the more recent date is considered more reliable.
\end{quote}

Given the current time specified in the system prompt and applying this rule, the Reasoner concludes that $C_2$ is the more credible source.
    
(4) \textbf{Final Synthesis:} The feedback loop confirms the resolution is sufficient. The model generates the final answer ($A_{final}$): ``\textit{Tesla's headquarters is located in Austin, Texas currently.}''

\begin{table}[h]
\centering
\small
\caption{Comparison of reasoning processes between Baseline and Our Method on the Tesla HQ query.}
\label{tab:case_study_process}
\begin{tabular}{p{0.15\linewidth} | p{0.38\linewidth} | p{0.38\linewidth}}
\toprule
\textbf{Stage} & \textbf{Baseline (Vanilla ICL)} & \textbf{\our} \\
\midrule
\textbf{Input} & Query + Context (Palo Alto, Austin) & Query + Context (Palo Alto, Austin) \\
\midrule
\textbf{Internal State} & High prob. for ``Palo Alto''. & ``HQ is in Palo Alto as of 2021.'' (Explicit text) \\
\midrule
\textbf{Conflict Handling} & \textit{Implicit Voting:} Internal weight > Context weight. & \textbf{Analyzer:} Flags contradiction $(C_i,C_j)$. \\
\midrule
\textbf{Resolution} & Ignored/Hallucinated Merge. & \textbf{Reasoner:} Applies \textit{Temporal Update Rule} (``more recent date is considered more reliable''). \\
\midrule
\textbf{Outcome} & \textcolor{red}{\textbf{Incorrect:}} Palo Alto. & \textcolor{green}{\textbf{Correct:}} Austin, Texas. \\
\bottomrule
\end{tabular}
\end{table}

\subsubsection{Generated Resolution Explanation}
Beyond the final answer, a key feature of \our is the production of an interpretable explanation $E$ that justifies the conflict resolution. For the Tesla scenario, the explanation generated by the Reasoner is shown below:

\begin{tcolorbox}[colback=gray!10, colframe=black!70, title=\textbf{Generated Conflict Resolution Explanation ($E$)}, arc=2mm]
\small
\textbf{[Conflict Identification]:} A factual discrepancy exists regarding the location of Tesla's headquarters. The context $C_0$ and $C_1$ support ``Palo Alto'', whereas context $C_2$ supports ``Austin''.\\
\textbf{[Resolution Logic]:} The conflict is categorized as a Temporal Conflict. The system applied Rule \#12 with recency principle. Analysis shows that $C_2$ explicitly states the location is current ``since 2021'', which post-dates the information in $C_1$ (describing a past move) and the temporal scope of $C_0$.\\
\textbf{[Conclusion]:} The information in $C_2$ is deemed the most credible source due to temporal precedence. The final answer is grounded in $C_2$, overriding the outdated consensus of $C_0$ and $C_1$.
\end{tcolorbox}

\section{Limitations}
While \our demonstrates significant improvements in accuracy, it presents two primary limitations. First, the efficiency is constrained by the multi-agent architecture; the iterative interactions between the Analyzer and Reasoner, along with the feedback loops, necessitate frequent LLM calls, increasing both latency and computational cost compared to single-pass methods. Second, the stability of the rule induction process relies heavily on the capabilities of the underlying LLM. The quality and generalizability of the induced rules can fluctuate depending on the model's ability to abstract patterns from demonstration data, leading to potential unpredictability in rule application for highly ambiguous. One promising direction is to integrate symbolic reasoning components and optimize inter‑agent communication, thereby improving efficiency and stability while preserving the interpretability of the overall framework.

\section{Conclusion}
\label{sec:conclusion}
This paper addresses the critical challenge of knowledge conflict in large language models, where discrepancies between a model’s parametric memory and retrieved contexts, or among the contexts themselves, can result in factual errors or hallucinations. Existing approaches typically assume that at least one source is reliable, directly adopt the chosen source as the final answer without exposing the reasoning process, thereby limiting interpretability.

To overcome these limitations, we propose a novel, interpretable framework that combines an adaptive knowledge assessment and retrieval approach with an inductive multi-agent reasoning paradigm. Our method explicitly extracts and evaluates the model’s internal knowledge alongside external contexts, allowing conflict resolution without presupposing which source is trustworthy. Through coordinated interaction among three agents, the system inductively learns and applies logical rules for resolving conflicts, producing comprehensive justifications.
Extensive experiments on the three benchmarks demonstrate that our framework significantly outperforms state-of-the-art baselines. 

Future work will focus on optimizing the computational efficiency of the multi-agent interaction to support real-time applications. Additionally, we aim to promote the combination of structured rules and parametric models, further advancing the reliability and trustworthiness of Large Language Models in complex information environments.
%
\bibliographystyle{IEEEtran}
\bibliography{sample-base}

\end{document}